\pdfoutput=1

\documentclass[11pt]{article}

\usepackage[]{EMNLP2022}

\usepackage{times}
\usepackage{latexsym}

\usepackage[T1]{fontenc}
\usepackage[utf8]{inputenc}
\usepackage[final]{microtype}
\usepackage{inconsolata}

\usepackage{multirow}

\usepackage[english]{babel}
\usepackage{scrlfile} 
\usepackage{amssymb}
\AfterPackage{amsmath}{\usepackage[nameinlink]{cleveref}}
\usepackage[final]{graphicx}
\usepackage{enumitem}
\usepackage{verbatim}
\usepackage{siunitx}
\usepackage{csquotes}
\usepackage{makecell}
\usepackage{booktabs}
\usepackage{tabularx}
\usepackage{tikz}

\usepackage[
	acronym,
	nomain, 
	nogroupskip,
	nonumberlist, 
	nopostdot,
	toc
]{glossaries-extra}

\setabbreviationstyle[acronym]{long-short}

\newacronym{nlp}{NLP}{natural language processing}

\newacronym{mt}{MT}{machine translation}
\newacronym{lm}{LM}{language model}
\newacronym{nnlm}{NN-LM}{nearest neighbour \acrlong{lm}}
\newacronym{knn}{kNN}{$k$-nearest neigbour}
\newacronym{knnlm}{kNN-LM}{\acrlong{knn} \acrlong{lm}}
\newacronym{knnmt}{kNN-MT}{\acrlong{knn} \acrlong{mt}}
\newacronym{rnn}{RNN}{Recurrent Neural Network}
\newacronym{trf}{TRF}{Transformer}

\newacronym{wt103}{WT103}{Wikitext-103}
\newacronym{spc}{SPC}{Stanford Politeness Corpus}
\newacronym{gyafc}{GYAFC}{Grammarly's Yahoo! Answers Formality Corpus}
\newacronym{rtp}{RTP}{Real Toxicity Prompts}
\newacronym{tox1}{TOX1}{\emph{Jigsaw Unintended Bias in Toxicity Classification} dataset}
\newacronym{tox2}{TOX2}{\emph{Real Toxicity Prompts} dataset}

\newacronym{mae}{MAE}{Mean Absolute Error}
\newacronym{mbe}{MBE}{Mean Bias Error}
\newacronym{iqr}{IQR}{Interquartile Range}
\newacronym{ppl}{%
PPL%
}{Perplexity}
\newacronym{sd}{SD}{Standard Deviation}

\newacronym{sota}{SOTA}{state of the art}
\newacronym{ir}{IR}{Information Retrieval}
\newacronym[shortplural={DSes}]{ds}{DS}{Datastore} 
\newacronym{bpe}{BPE}{Byte Pair Encoding}
\newacronym{amt}{AMT}{Amazon Mechanical Turker}

\newglossaryentry{sbm}{
  name			= {sample break mode},
  type			= acronym,
  description	= {}
}

\setkeys{glslink}{hyper=false} 

\title{Nearest Neighbor Language Models for Stylistic Controllable Generation}

\author{Severino Trotta $\dagger$ \and Lucie Flek $\dagger \ddagger$ \and Charles Welch $\dagger \ddagger$\\
    $\dagger$ Conversational AI and Social Analytics (CAISA) Lab \\ 
    Department of Mathematics and Computer Science, University of Marburg \\ 
    $\ddagger$ The Hessian Center for Artificial Intelligence (Hessian.AI) \\
    \texttt{severino.trotta@gmail.com, \{lucie.flek,welchc\}@uni-marburg.de} 
} 

\begin{document}

\newcolumntype{L}[1]{p{\dimexpr #1-2\tabcolsep}}
\newcolumntype{R}[1]{>{\raggedleft\arraybackslash}p{\dimexpr #1-2\tabcolsep}}
\newcolumntype{P}[1]{>{\centering\arraybackslash}p{\dimexpr #1-2\tabcolsep}}
\newcolumntype{M}[1]{>{\centering\arraybackslash}m{\dimexpr #1-2\tabcolsep}}

\maketitle

\begin{abstract}
Recent language modeling performance has been greatly improved by the use of external memory. This memory encodes the context so that similar contexts can be recalled during decoding. This similarity depends on how the model learns to encode context, which can be altered to include other attributes, such as style. We construct and evaluate an architecture for this purpose, using corpora annotated for politeness, formality, and toxicity. Through extensive experiments and human evaluation we demonstrate the potential of our method to generate text while controlling style. We find that style-specific datastores improve generation performance, though results vary greatly across styles, and the effect of pretraining data and specific styles should be explored in future work.
\end{abstract}

\section{Introduction}

Language models with external memory, like \citet{knnlm}'s recent \gls{knnlm}, have demonstrated impressive predictive performance. Great reductions in perplexity are achieved through storing the encoding of contexts from the training data. A sequence of tokens is encoded by the model and stored as a key, which is paired with a value representing the next token in the sequence. During decoding, similar contexts are recalled based on their key similarity, and values are interpolated with the base \glsfmtlong{lm}'s predictions.

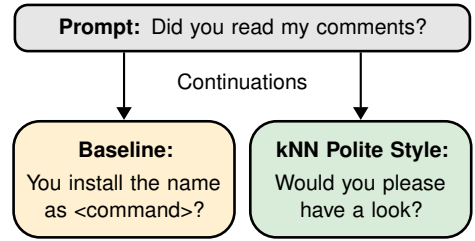
\begin{figure}
    \centering
    \usetikzlibrary{positioning, fit, calc, arrows.meta}
    \begin{tikzpicture}[
        font=\sffamily\fontsize{8}{10}\selectfont, node distance=0.8ex, >={Triangle},
        promptbox/.style={
            draw=black, line width=0.2ex, rounded corners=1ex, text width=0.8\linewidth-4ex, align=center, inner ysep=1ex, inner xsep=1.5ex
        },
        contbox/.style={
            promptbox, rounded corners=2ex, text width=0.4\linewidth-3ex, minimum height=4em, inner ysep=1ex, inner xsep=1ex
        }
    ]
        \node[contbox, fill=yellow!50!orange!20] (cont-baseline) {\textbf{Baseline:}\\[0.6ex] You install the name as <command>?};
        \node[right=1ex of cont-baseline, contbox, fill=green!50!black!15] (cont-polite) {\textbf{kNN Polite Style:}\\[0.6ex] Would you please have a look?};
        \node[fit={(cont-baseline) (cont-polite)}] (conts) {};

        \node[above=of conts, draw=none] (cont-label) {Continuations};
        \node[above=of cont-label, promptbox, fill=black!10] (prompt) {\textbf{Prompt:} Did you read my comments?};
        \draw[->, thick] (prompt.south -| cont-baseline.north) -| (cont-baseline.north);
        \draw[->, thick] (prompt.south -| cont-polite.north) -| (cont-polite.north);
    \end{tikzpicture}
    \caption{Example prompt continuations of the baseline \glsfmtshort{knn} \glsfmtlong{lm} and our model, which continues generation in a specified style (e.g. polite). This example is based on a real example from our human evaluation but shortened for brevity and clarity. Full examples are provided in \Cref{sec:appendix-full-examples}.}
    \label{fig:knn_example}
\end{figure}

In this work, we augment the encoding with stylistic attributes, such that the keys are more heavily influenced by the style of the encoded text. By explicitly encoding the style, the similarity is more strongly affected by the stylistic attributes than previous models. When decoding, we can then provide a style (e.g. polite or formal) and the most similar contexts are both relevant in content and more likely to conform to the provided style. The example in \Cref{fig:knn_example} shows a prompt and two continuations, one with the baseline \glsfmtshort{knn} \glsfmtlong{lm} and one with our model given a polite style value as input, signaling that it should continue the prompt in a polite style.

Through our architecture implementation, we show that we not only improve language modeling performance over previous models, but that we can control generation to produce text of a particular style. We provide human evaluation of our stylistic outputs and an analysis of the performance of our approach and modeling decisions that affect how style attributes are represented in memory. To the best of our knowledge, this is the first work to modify a \glsfmtlong{lm}'s external memory in order to control generated style.

\section{Related Work} \label{sec:related-work}

Recent approaches to controllable generation have included fine-tuning large models, such as \citet{keskar2019ctrl}, who condition on \num{50} control codes during training, which represent different styles, topics, and languages. Other approaches avoid retraining by modifying the predictions only at decoding time. The FUDGE model predicts for a sequence, the likelihood that possible generation steps will result in a sequence that satisfies a given constraint \citep{yang-klein-2021-fudge}. The DExperts model alters the probability distribution of a \gls{lm} based on the predictions of other \glspl{lm} that are fine-tuned on specific desired or undesired attributes \citep{liu-etal-2021-dexperts}. \citet{dathathri2019plug}  and others similarly modify gradients directly during prediction.

\citeauthor{knnlm}'s \glsfmtlong{knnlm} is based on the groundwork laid by previous work that augmented language models with a cache memory of recent observations. \citet{continuous-cache} captured local context of up to a few thousands tokens to improve predictions. \citet{unbounded-cache} expanded upon this idea by storing all past hidden activations in a memory. \citeauthor{knnlm} then replaced the recurrent network with a transformer to better model long-term dependencies. Aspects of \glspl{knnlm} have been improved upon, in terms of performance, efficiency, and additional functionality. 

Even though \glspl{knnlm} achieve state-of-the-art predictive performance, the retrieval operation is very computationally expensive. 
\citet{efficient-knnlm} and \citet{automaton-augmented-retrieval} explored techniques to improve inference speed, such as compressing embeddings, or training an additional model to dynamically disable retrieval for predictions where the datastore is unlikely to improve the result. \citet{wu2022memorizing} implemented a similar model and focused on improving scalability. \citet{yogatama-etal-2021-adaptive} extended the model with a gating mechanism that learns to combine short and long term memory with local context. \citet{structured-locality} improved performance by leveraging structural locality features such as topic clusters in text or project hierarchies in source code repositories. \citet{knnlm} also extended their model for use in machine translation \citep{knn-mt}, which has also received efficiency improvements \citep{knn-mt-fast, knn-mt-faster}.

\section{Methodology}\label{sec:methodology}
The main goal of our work is to expand \gls{knnlm} functionality and we build off of \citet{knnlm}, which we will refer to as the baseline architecture.
We modified this architecture to accept additional style attributes as input, and concatenate these to the input text encoding. This has the effect of modifying the embedding space such that it encodes both semantic and stylistic properties (see \Cref{sec:appendix_arch}). We will refer to this as the style architecture.

After the input style and context are encoded, they are stored in the datastore.
We experimented with separating datastores based on the distribution of style values in the dataset. 
For this part we take subsets with specific style values 
(e.g. only toxic or only polite) 
from the datasets and construct datastores containing only data from those subsets. 
We refer to this as \emph{separate datastores}. Datastores containing examples of different style are referred to as \emph{mixed datastores}.

\section{Datasets}\label{sec:resources}

We use \num{4} datasets, \num{3} of which contain style attributes. Unless stated otherwise, we use the default splits provided by the original work.

\paragraph{Wikitext-103}
\gls{wt103} is a collection of \emph{Good} and \emph{Featured} Wikipedia articles \cite{merity2016pointer}. It is provided in a tokenized form, with case, punctuation and numbers, totalling \SI{103}{B} tokens. 
For better compatibility with our tokenization of other datasets we modified \gls{wt103}'s tokenization by replacing all occurrences of \verb|*n 't| (e.g. in \enquote{can't}) with \verb|* n't|.

\paragraph{Politeness}
The \gls{spc} consists of \si{11\kilo{}} utterances from StackExchange and Wikipedia Talk pages, annotated with politeness scores, which we use as style attributes \cite{danescu-politeness-2013}. We created an 85-7.5-7.5 split for training, validation, and test.

\paragraph{Formality}
\gls{gyafc} is the largest available corpus for formality style transfer, containing about 110K formal/informal sentence pairs \cite{gyafc-dataset}. It is divided into the domains \emph{Entertainment \& Music} and \emph{Family \& Relationships}, which makes it suitable for training and evaluation with in/out-of-domain data. We do not use the parallel nature of the corpus, but assign each sentence a style attribute (\num{-1} for informal- and \num{1} for formal sentences) and re-split the training subset to obtain an 80-10-10 train\slash{}validation\slash{}test split.

\paragraph{Toxicity}
For toxicity we use the Jigsaw Unintended Bias in Toxicity Classification dataset, which contains comments from the Civil Comments platform annotated with several binary toxicity labels representing types of toxicity (e.g. insult, identity attack, sexually explicit) \cite{toxicity-dataset}. We use only the \textit{toxic} label as a style attribute. We further use the Real Toxicity Prompts dataset for human evaluation \cite{toxicityprompts}. This dataset contains the beginning of sentences and has been used to test if models can continue generation without toxicity.

\subsection{Preprocessing}
We largely follow \citet{merity2016pointer}'s preprocessing of \gls{wt103} for all data. Additionally, we perform the following replacements:

\begin{center}
\small
\begin{tabular}{rl}
	\thead[r]{what} & \thead[l]{replacement} \\
	inline/block code & \verb|<code>| \\
	usernames & \verb|<person>| \\
	hyperlinks & \verb|[title](url)| $\mapsto$ title \\
	URLs & \verb|<url>| \\
\end{tabular}	
\end{center}

Unlike \citet{merity2016pointer} we use \verb|spaCy.io| 
for normalization and tokenization, but perform the same tokenization of infix punctuation and symbols. 
This serves to differentiate number separator punctuation from word punctuation, and hyphens from minus signs or ranges.

\section{Experiments}

In preliminary experiments we examined the effect of float precision on model performance.
Using \gls{wt103} and the setup of \citet{knnlm} we found that using half precision halves inference time without hurting perplexity, and reduces inference time for the \gls{knnlm} with a negligible \num{0.25} increase in perplexity. In the following sections we use half-precision to speed up the experimentation process. 

In the main experiments we first examine the impact on perplexity when incorporating style values into the model. Then we compare our method to the baseline through human evaluation.

\subsection{Language Modeling with Style Attributes}

We train the style model, $S$, using the modified architecture, and baseline model, $B$, which uses the original architecture.
To achieve high predictive performance on text, we first train both on \gls{wt103}. 

For each dataset we then fine-tune a copy of $S$ and $B$ on the dataset. 
Using these fine-tuned models we generate and evaluate a datastore for each a model, using the dataset the respective model was fine-tuned on.

We additionally build and evaluate a datastore on domain and style subsets, to test our model's performance on out-of-domain data and across subsets with specific style. 
The subsets are listed in \Cref{tbl:dataset-subsets} in the Appendix.

\subsection{Setup \& parameters}
Unless stated otherwise, we use the default parameters used by the \verb|transformer_lm_wiki103| architecture and \citet{knnlm}.

\paragraph{Vocabulary}
To avoid a high number of OoV tokens, we chose to use a shared vocabulary from the union of all datasets. 
Tokens occurring less than \num{3} times were mapped to \verb|<unk>|. 
The resulting vocabulary has a size of \si{375\kilo} tokens.
Since \citet{merity2016pointer} have shown that a large vocabulary with adaptive input representation can outperform a smaller BPE vocabulary, we chose the former over the latter -- although this choice has its own limitations (see \hyperref[sec:limitations]{limitations section}).

\paragraph{Training} During pretraining on \gls{wt103}, we use random style values normally distributed around the median, set patience to \num{5} epochs, and the style embedding dimension to \num{96}. During fine-tuning we use Adam instead of the NAG optimizer and set patience to \num{10} epochs.

\paragraph{Datastore} We use half-precision vectors in the datastore. Since the context embedding dimension (\num{1120}) must be divisible by the number of FAISS index subquantizers, we use \num{70} instead of \num{64}. For smaller subsets of data we use \num{2048} cluster centroids, rather than the default \num{4096}.

\subsection{Fine-tuning Results}
The results of fine-tuning for each style attribute are shown in \Cref{tab:fine-tune}. We find that simply encoding the style value improves model performance on all datasets, including the original \gls{wt103}, though this is likely due to the small increase in the number of parameters. 
Fine-tuning predictably lowers perplexity on the style datasets and slightly increases the \gls{wt103} perplexity as the model shifts away from the corpus it was originally trained on. The best performance is on formality, followed by politeness, which we expect to more closely resemble \gls{wt103}. The addition of the style input allows for much greater improvement on politeness as compared to toxicity which shows near equal performance without it. We also provide perplexity for the \gls{knnlm} in \Cref{sec:appendix-wiki-finetune}.

\begin{table}[ht]
	\centering
	\small
	\begin{tabular}{r*{4}{S[table-format=3]}}
		\toprule
		 & \multicolumn{2}{c}{{\thead{Baseline}}} & \multicolumn{2}{c}{{\thead{Style}}} \\
		\cmidrule(l{0pt}r{2pt}){2-3}
		\cmidrule(l{2pt}r{0pt}){4-5}
		\thead{Dataset} & {\thead{PT}} & {\thead{FT}} & {\thead{PT}} & {\thead{FT}} \\
		\midrule
		Politeness & 218 & 126 & 164 & 78 \\
		Formality & 161 & 77 & 148 & 60 \\
		Toxicity & 212 & 125 & 186 & 93 \\
		\midrule
		\gls{wt103} & 31 & {35-64} & 29 & {32-59} \\
		\bottomrule
	\end{tabular}
	\caption{%
		Perplexity before (using the pretrained model; PT) and after fine-tuning (FT). All models were evaluated on the FT dataset and on \gls{wt103}.
        The value ranges in the \gls{wt103} row indicate the performance range of the FT models on the \gls{wt103} dataset.
	}
    \label{tab:fine-tune}
\end{table}

\subsection{Human Evaluation}\label{sec:human_eval}

We aimed to answer three questions; (1) do the style-specific datastores outperform the mixed datastore, (2) does the \gls{knnlm} outperform the \gls{lm}, and (3) does the style architecture outperform the baseline?
To address these questions we asked a group of \num{11} students to annotate model outputs.

\paragraph{Generating Output}
We follow the idea of \citet{toxicityprompts} and generate outputs by supplying prompts of different styles to the models. We use both non-toxic (toxicity $=\num{0}$) and neutral ($ 0 < \text{toxicity} < \num{0.5}$) prompts from the toxicity dataset, and created prompts from the formality and politeness datasets by cutting off the second half of randomly sampled sentences. The prompts are then used as input to the models, which generate continuations to the prompts. All combinations of models and inputs are detailed in \Cref{tbl:finetune-prompts-model-combinations} in the Appendix. 
For all \gls{knnlm} outputs we use $\lambda=\num{0.8}$ as interpolation parameter.

\paragraph{Survey Setup}
We asked annotators to select which of a pair of prompt continuations was more fluent and which more closely followed one of the given styles.
The pair combinations are based on the three comparisons listed at the beginning of this section, but are fully listed in \Crefrange{tbl:survey-output-pairs-part1}{tbl:survey-output-pairs-part3} in the Appendix and result in a total of \num{440} survey questions.
The questions were presented to annotators in random subsets of \SI{20}{\%} of the full set.  
Each output pair was rated by \numrange{2}{4} people.

\paragraph{Results}

\begin{table}
    \centering
    \sisetup{
		round-mode=places, 
		round-precision=1
    }
    \small
    \begin{tabular}{rc S[table-format=2.1] S[table-format=2.1]}
        \toprule
        & & {\thead{Fluency (\%)}} & {\thead{Style (\%)}} \\
        \midrule
        \multirow{2}{*}{Datastore} & Mixed & 45.833333 & 48.194444 \\
        & Specific & 54.1667 & 51.8056 \\
        \midrule
        \multirow{2}{*}{Model type} & \glsfmtshort{lm} & 47.708333 & 50.694444 \\
        & \glsfmtshort{knnlm} & 52.2917 & 49.3056 \\
        \midrule
        \multirow{2}{*}{Architecture} & Baseline & 48.791667 & 47.291667 \\
        & Style & 51.2083 & 52.7083 \\
        \bottomrule
    \end{tabular}
    \caption{
    Human evaluation preferences for model pairs. Column-wise percentage pairs sum to \num{100}.
    }
    \label{tab:human_evaulation}
\end{table}

The results in \Cref{tab:human_evaulation} show which model is preferred, in percentage of annotators, in terms of fluency and style for each pair. We find that when comparing mixed and specific datastores, the specific datastores are preferred for style and even more strongly for fluency. While the \gls{knnlm} is preferred over the \gls{lm} in fluency, the style preference is more evenly split. When comparing the style architecture to the baseline, we find that the ours is preferred, with style more strongly preferred to fluency.

These results are an aggregation over the styles, however the performance on specific styles reveals more varied results. The specific datastores give more style control for politeness than for other styles and for some combinations of the prompt style and target style, the mixed datastore was preferred. We see that when we want to generate non-toxic, polite, or formal text; those that more closely resemble the pretraining data style, the preference leans more toward the mixed datastore.

When comparing the \gls{lm} to the \gls{knnlm}, we found that the \gls{lm} style was often preferred when provided an informal, non-toxic, or impolite prompt, regardless of the target style. We also found that the fluency of the \gls{lm} is preferred when generating polite or impolite text. Lastly, the style architecture is not always preferred over the baseline either. The baseline shows stronger fluency for toxic and impolite prompts. The style architecture has the best style control when generating formal, toxic, and impolite text. Overall, there appears to be a trade-off between style-control and fluency. The full breakdown by prompt and input type is shown in \Crefrange{fig:finetune-human-evaluation-part1}{fig:finetune-human-evaluation-part3} in the appendix.

\section{Conclusion}

We examined the use of \glsfmtshort{knn} \glsfmtlongpl{lm} for controllable stylistic generation using politeness, formality, and toxicity as target styles. Our findings show that simply encoding style in the architecture improved perplexity of the \glsfmtlong{lm}. A human evaluation further showed that specific datastores for target styles outperform the standard mixed datastore, and that our model generally outperformed the baseline \glsfmtshort{knn} model in terms of fluency and style control, though results on specific styles varied. Future work is needed to fully understand the effect of pretraining and benefits of the model variants for specific styles and should also consider comparisons to other controllable generation models, such as \citet{keskar2019ctrl}. 

Our code is available on Github at \url{https://github.com/d8xa/style-knnlm}.

\section*{Limitations}\label{sec:limitations}

\paragraph{Vocabulary Choice}
We chose a shared vocabulary to reuse the same baseline model for fine-tuning on multiple datasets. Since less frequent tokens are assigned less parameters by the adaptive input representation, this could lead to under-representation of rare, style-specific tokens in general, and worse fine-tuning results for smaller datasets or datasets with many rare tokens. The same problem applies for single-dataset vocabularies as well, when rare tokens are more prevalent for a particular style. Byte-pair encoding would avoid these problems, but make comparability to the vanilla \gls{knnlm} more difficult.

\paragraph{Sequence Length}
The original \gls{knnlm} was trained with sequences of up to \num{3072} tokens in length, which helps model long-term dependencies in the \gls{wt103} dataset. Since all of our datasets with style attributes contain much shorter sequences, single-dataset training with shorter input sizes might be better suited and achieve better performance than pre-training on \gls{wt103} and fine-tuning on the style dataset.

\paragraph{Comparability with \citet{knnlm}}
When training our style architecture, we had to choose between a combined embedding dimension of $C+S_{\text{emb}}=\num{1024}$ (token- and style embedding dimensions $C$ and $S_{\text{emb}}$), or to use $C=\num{1024}$. In any case the resulting \glsfmtlong{lm} would have a different number of parameters than in the original \gls{knnlm} . We chose to use $C=\num{1024}$ and $S_{\text{emb}}=\num{96}$. FAISS requires the vector dimension to be divisible by the number of subquantizers. Since our combined embedding dimension is different from \num{1024}, we had to choose \num{70} instead of \num{64} subquantizers.

Another difference is the choice of vocabulary. The \gls{wt103}-only vocabulary would make results more comparable, but also lead to a high number of \verb|UNK| tokens for the style datasets, and therefore reduce performance greatly.

\paragraph{Token to Style Embedding Dimension Ratio}
To limit the scope of this work we did not perform an analysis on the ratio between token- and style embedding dimension. Other ratios might achieve better fluency or style control.

\paragraph{Choice of Interpolation Parameter}
For our human evaluation we put $\lambda=\num{0.8}$ weight on \gls{knnlm} probabilities. 
A lower $\lambda$, more close to the vanilla \gls{knnlm}, might achieve better fluency.

\section*{Ethics Statement}

Work on controllable generation allows models to generate text in styles such as those presented here, which include polite\slash{}impolite, formal\slash{}informal, and toxic\slash{}non-toxic. There may be applications where each style is appropriate or desirable, but some styles of text such as impolite and toxic, may be undesired or even harmful. Application of our models should involve careful consideration of desired styles and the context in which they are deployed.

\section*{Acknowledgements}
This work has been supported by the Alexander von Humboldt Foundation, and by Hessian.AI. Any opinions, findings, conclusions, or recommendations in this material are those of the authors and do not necessarily reflect the views of the Alexander von Humboldt Foundation, or Hessian.AI. 

\bibliography{references}
\bibliographystyle{acl_natbib}

\appendix

\section{Generation Examples}\label{sec:appendix-full-examples}

Here we include three examples of generation comparing the baseline \gls{lm} to the \gls{knnlm} where human annotators found our model to be both more fluent and more closely aligned with the desired target style.

\begin{table}[ht!]
    \centering
    \begin{tabularx}{\linewidth}{rX}
    \toprule
    \multicolumn{2}{c}{\thead{Target Style: Polite}} \\
    \midrule
        \textbf{Prompt:} & <person>: Yes. Did you read my comments? \\
        \textbf{Baseline:} & Also, you install the name as <unk> ('') <url>?? \\
        \textbf{Our Model:} & Vote under the changed circumstances? Would you please have a look? \\
    \midrule
    \multicolumn{2}{c}{\thead{Target Style: Formal}} \\
    \midrule
        \textbf{Prompt:} & ok let me answer u \\
        \textbf{Baseline:} & 'm not in pain, and I say ... graham understand your problem \\
        \textbf{Our Model:} & are alike. You should wait a little longer to enter your full name and play your live show. \\
    \midrule
    \multicolumn{2}{c}{\thead{Target Style: Formal}} \\
    \midrule
        \textbf{Prompt:} & THis is one way to keep it out of your system ... \\
        \textbf{Baseline:} & sings u love bio <unk> \\
        \textbf{Our Model:} & with the following words coming out of Gardisil ... Peter Pan, or just those are some favorite books that I am able to mail. \\
        \bottomrule
    \end{tabularx}
\end{table}

\section{Modified Architecture}\label{sec:appendix_arch}
\begin{figure}
	\centering
	\resizebox{\linewidth}{!}{
	\includegraphics[height=0.3\textheight]{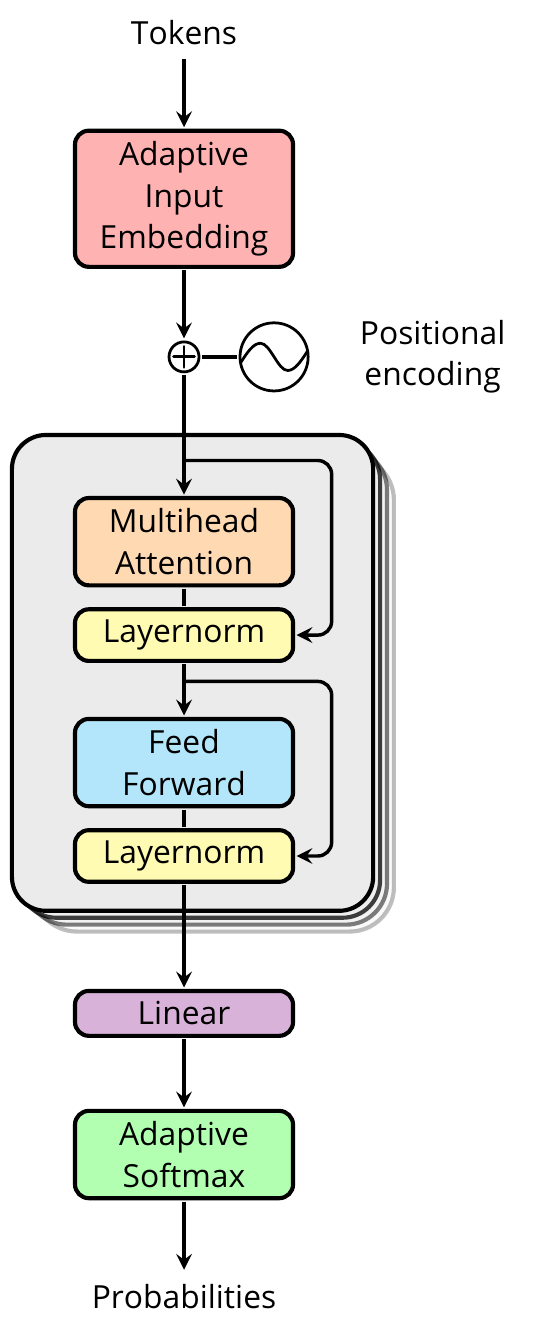}
	\includegraphics[height=0.3\textheight]{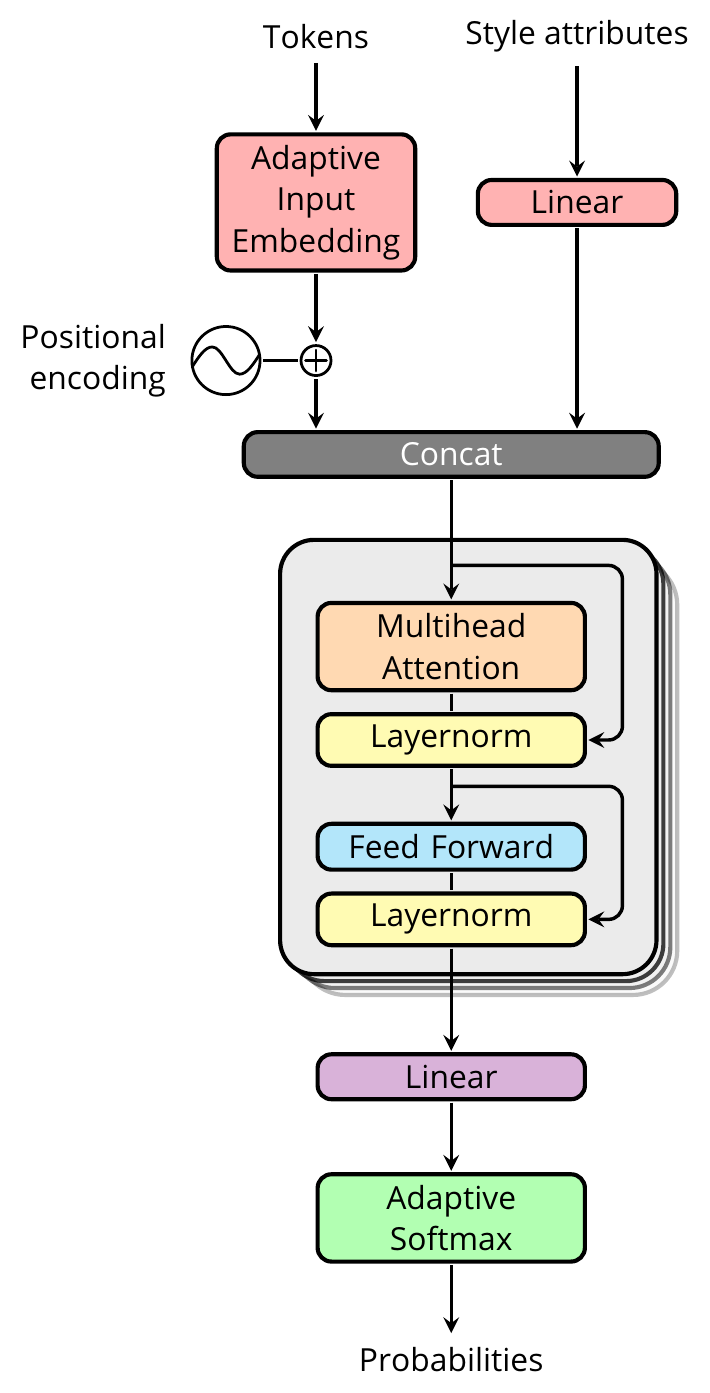}
	}
	\caption{
		Changes made to the \gls{lm} architecture
		(left: unmodified, right: our version).
	}
	\label{fig:arch}
\end{figure}

Our architecture, modified from \citet{knnlm}, is shown in \Cref{fig:arch}. The left side of the figure shows the original model and the right shows our modification which incorporates style attributes.

\begin{table*}[ht]
	\centering
	\small
	\sisetup{
		round-mode=places, 
		round-precision=1,
		zero-decimal-to-integer,
	}
	\begin{tabular}{l R{0.25\linewidth} S[
		zero-decimal-to-integer
		] c L{0.35\linewidth}}
		\toprule
		\thead[l]{dataset} & \thead[r]{subset} & \multicolumn{2}{c}{{\thead[c]{size}}} & \thead[l]{comment} \\
		\cmidrule(l{1pt}r{1pt}){3-4}
		& & {\thead[c]{samples (\%)}} & \thead[c]{samples} & \\
		\midrule
		\gls{wt103} 
			& all & 100 & \SI{1.809468}{\mega{}} & \\
		\midrule
		\multirow{5}{*}{Formality} 
			& all & 100 & \SI{213.973}{\kilo{}} & \\
			& formal & 49.84881 & \SI{106.663}{\kilo{}} & \\
			& informal & 50.15119 & \SI{107.310}{\kilo{}} & \\
			& family \& relationships & 49.67215 & \SI{106.285}{\kilo{}} & \\
			& entertainment \& music & 50.32785 & \SI{107.688}{\kilo{}} & \\
		\midrule
		\multirow{7}{*}{Toxicity} 
			& all & & \SI{1.358102}{\mega{}} & Only used for evaluation; due to imbalance not suited for training or \glspl{ds}. \\
			& non-toxic & 75.6075022 & \SI{1.026827}{\mega{}} & Toxicity score $=\num{0}$ .\\
			& non-toxic-sample & 11.7650957 & \SI{159.782}{\kilo{}} & Sample from \emph{non-toxic}. \\
			& toxic-gte-0.5 & 11.7650957 & \SI{159.782}{\kilo{}} & Toxicity score $\geq \num[round-precision=1]{0.5}$. \\
			& toxic-gte-0.8 & 2.5136551 & \SI{34.138}{\kilo{}} & Toxicity score $\geq \num[round-precision=1]{0.8}$. \\
			& toxic-gte-0.9 & 0.7528153 & \SI{10.224}{\kilo{}} & Toxicity score $\geq \num[round-precision=1]{0.9}$. \\
			& all-sample & 23.5301914 & \SI{319.564}{\kilo{}} & Combination of \emph{toxic-gte-0.5} and \emph{non-toxic-sample}. \\
		\midrule
		\multirow{6}{*}{Politeness}
			& all & 100 & \SI{11.118}{\kilo{}} & \\
			& neutral & 30.32920 & \SI{3.372}{\kilo{}} & Center \SI{30}{\%} of politeness scores. \\
			& polite & 36.90412 & \SI{4.103}{\kilo{}} & Upper \SI{36.90412}{\%} of politeness scores. \\
			& impolite & 32.75769 & \SI{3.642}{\kilo{}} & Lower \SI{32.75769}{\%} of politeness scores. \\
			& stackexchange & 60.84727 & \SI{6.765}{\kilo{}} & \\		
			& wikipedia & 39.15273 & \SI{4.353}{\kilo{}} & \\
		\bottomrule
	\end{tabular}
	\caption{
		List of dataset subsets.
		Note: Proportions of subsets within splits are subject to variations due to random sampling. Not all subsets are presented in the results of the main paper. Some subsets are only shown in \Cref{fig:wiki-finetune-ppl}.
	}
	\label{tbl:dataset-subsets}
\end{table*}

\section{Fine-tuning Experiment}\label{sec:appendix-wiki-finetune}

The fine-tuning experiments in the main paper summarize the performance of our models pretrained on \gls{wt103} and fine-tuned on one of the style datasets each. Here we also include the performance on other subsets of the finetuning dataset, such as different toxicity levels for the toxicity data, and domain subsets for the politeness and formality data. A full list of subsets is given in \Cref{tbl:dataset-subsets}.

\begin{figure*}
	\centering
	\includegraphics[width=1.0\textwidth]{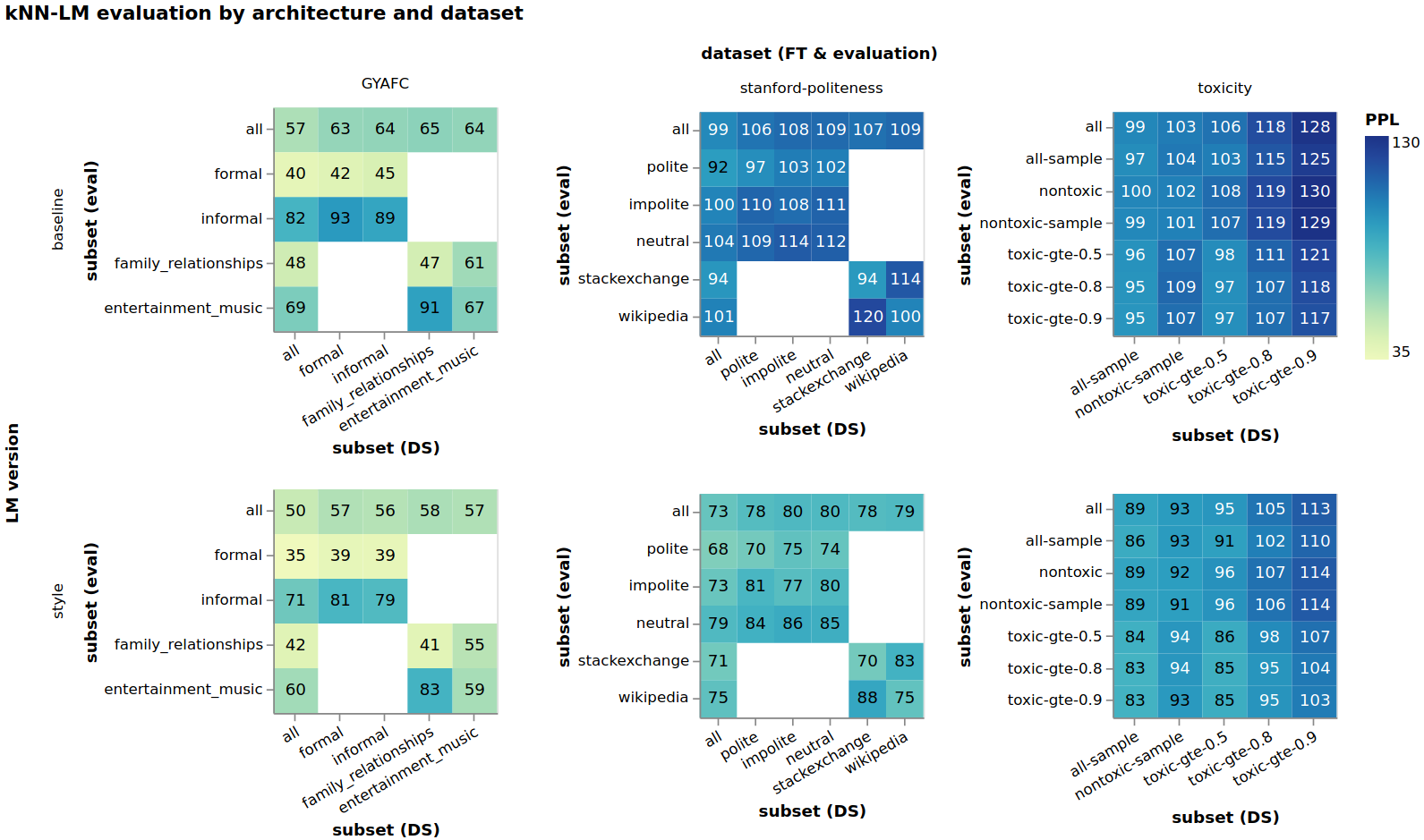}
	\caption{
		Overview of test perplexity in the fine-tuning experiment across data subsets listed in \Cref{tbl:dataset-subsets}.
	}
	\label{fig:wiki-finetune-ppl}
\end{figure*}

\section{Human Evaluation}

\begin{figure*}[ht]
	\centering
	\includegraphics[width=1.0\linewidth]{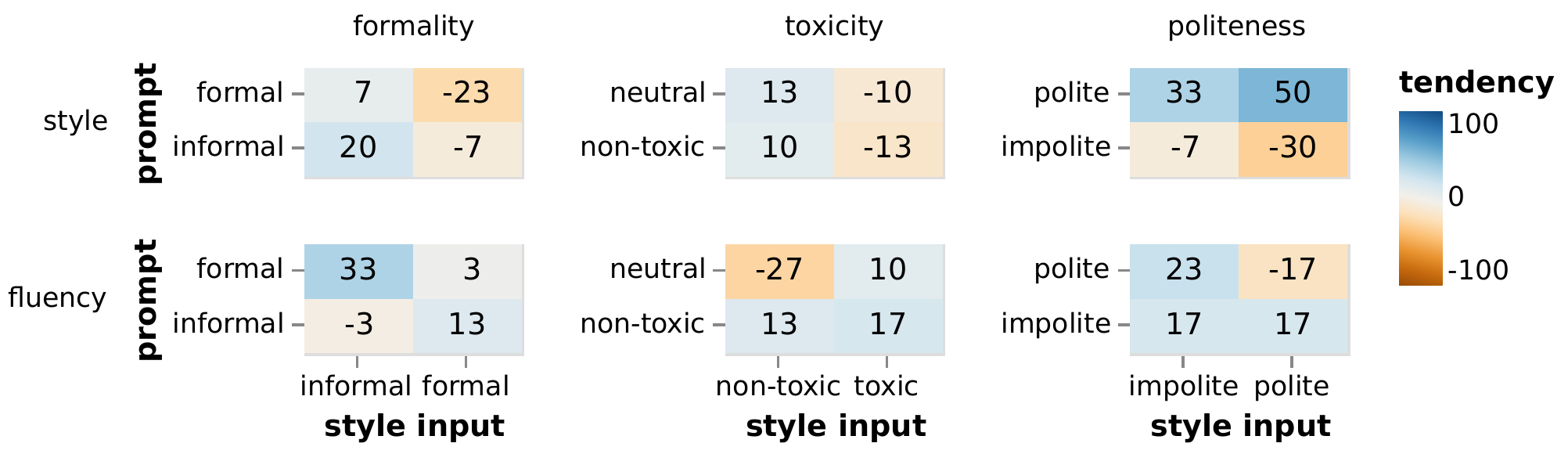}
	\caption{
		Survey results: Comparison of \gls{knnlm} with mixed \gls{ds} and style-specific \gls{ds}. 
        A tendency $<0$ corresponds to the mixed \gls{ds} being preferred by annotators.
	}
	\label{fig:finetune-human-evaluation-part1}
\end{figure*}

\begin{figure*}[ht]
	\centering
	\includegraphics[width=1.0\linewidth]{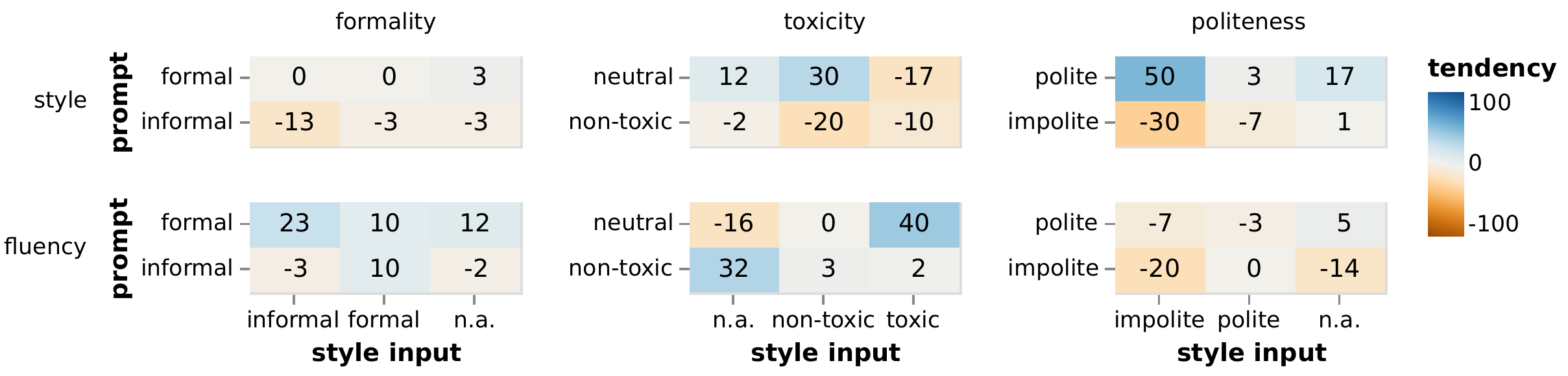}
	\caption{
		Survey results: Comparison of \gls{lm} to \gls{knnlm} for both architectures. A tendency $<0$ corresponds to the \gls{lm} being preferred by annotators.
	}
	\label{fig:finetune-human-evaluation-part2}
\end{figure*}

\begin{figure*}[ht]
	\centering
	\includegraphics[width=1.0\linewidth]{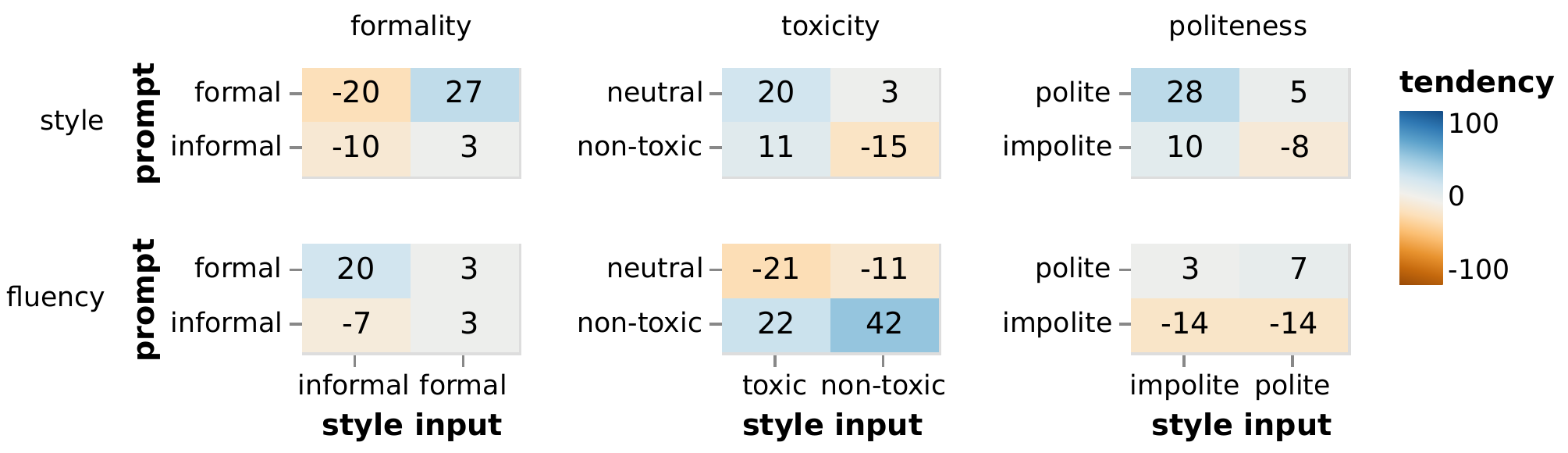}
	\caption{
		Survey results: Comparison of \gls{knnlm} with baseline architecture vs. style architecture. 
		Style input on the $x$-axis refers only to the style \gls{lm}, since the baseline \gls{lm} has no style input.
		A tendency $<0$ corresponds to the the baseline architecture being preferred by annotators.
	}
	\label{fig:finetune-human-evaluation-part3}
\end{figure*}

For the human evaluation task, we generated sentences for different combinations of the model architecture, target style, and prompt style. The high-level summary of combinations is presented in \Cref{tbl:finetune-prompts-model-combinations}. For the three test conditions in \S\ref{sec:human_eval}, we have listed the model and style combinations that we tested in \Crefrange{tbl:survey-output-pairs-part1}{tbl:survey-output-pairs-part3}. Finally, the full breakdown of the human evaluation for fluency and style preferences are shown in \Crefrange{fig:finetune-human-evaluation-part1}{fig:finetune-human-evaluation-part3}. The heatmaps show the tendency of annotator choices, where tendency is the mean model choice per question (\num{-1} and \num{1} encode the choices), aggregated across all questions in the survey and normalized to \num{-100} and \num{100}.

\begin{table*}
	\centering
	\small
	\begin{tabular}{lllll}
		\toprule
		\thead[l]{dataset} & \thead[l]{source style} & \thead[l]{target style} & \thead[l]{LM} & \thead[l]{datastore styles} \\
		\midrule
		Formality &      formal & formal & style &            none, formal, mixed \\
		      &        		& informal & style &          none, informal, mixed \\
  		      &        		& n.a. & baseline &         none, informal, formal \\
		      &    informal & formal & style &            none, formal, mixed \\
		      &        		& informal & style &          none, informal, mixed \\
		      &   		    & n.a. & baseline &         none, formal, informal \\
            \midrule
		  Toxicity & neutral & n.a. & baseline &  none, non-toxic, toxic, mixed \\
		      &        		& non-toxic & style &         none, non-toxic, mixed \\
		      &        		& toxic & style &             none, toxic, mixed \\
		      &   non-toxic & n.a. & baseline &  none, non-toxic, toxic, mixed \\
		      &        		& non-toxic & style &         non-toxic, none, mixed \\
		      &   		    & toxic & style &             none, toxic, mixed \\
		  \midrule
		  Politeness &    impolite & impolite & style &          none, impolite, mixed \\
		      &        		& n.a. & baseline &  none, polite, impolite, mixed \\
		      &        		& polite & style &            none, polite, mixed \\
		      &      polite & impolite & style &          none, impolite, mixed \\
		      &        		& n.a. & baseline &  none, polite, impolite, mixed \\
		      &        		& polite & style &            none, polite, mixed \\
		\bottomrule
	\end{tabular}
	\caption{
		Combinations of models and inputs used for generating the outputs for human evaluation. 
		\emph{n.a.} in the \emph{target style} column refers to the baseline \gls{lm} architecture, since it has no style input.
	}
	\label{tbl:finetune-prompts-model-combinations}
\end{table*}

\begin{table*}
	\centering
	\small
	\begin{tabular}{llllll}
		\toprule
		\thead[l]{dataset} & \thead[l]{source style} & \thead[l]{target style}  & \thead[l]{specific datastore style} \\
		\midrule
		Formality &   formal    & formal    & formal \\
		      &             & informal  & informal \\
		      &   informal  & formal    & formal \\
		      &             & informal  & informal \\
		      \midrule
		Toxicity   &   neutral   & non-toxic & non-toxic \\
		      &             & toxic     & toxic \\
		      &   non-toxic & non-toxic & non-toxic \\
		      &             & toxic     & toxic \\
		      \midrule
		Politeness   &   impolite  & impolite  & impolite \\
		      &             & polite    & polite \\
		      &   polite    & impolite  & impolite \\
		      &             & polite    & polite \\
		\bottomrule		                
	\end{tabular}
	\caption[Survey setup: Model output combinations (Part \num{1})]{
		Model combinations for the human evaluation to address whether the style-specific datastores outperform the mixed datastores.
	}
	\label{tbl:survey-output-pairs-part1}
\end{table*}

\begin{table*}
	\centering
	\small
	\begin{tabular}{llll}
		\toprule
		\thead[l]{dataset} & \thead[l]{source style} & \thead[l]{target style}   & \thead[l]{datastore style} \\
		\midrule
		Formality   & formal    & formal     & formal    \\
		        &           & informal   & informal  \\
		        & informal  & formal     & formal    \\
		        &           & informal   & informal  \\
		        \midrule
		Toxicity     & neutral   & non-toxic  & non-toxic \\
		        &           & toxic      & toxic     \\
		        & non-toxic & non-toxic  & non-toxic \\
		        &           & toxic      & toxic     \\
		        \midrule
		Politeness     & impolite  & impolite   & impolite  \\
		        &           & polite     & polite    \\
		        & polite    & impolite   & impolite  \\
		        &           & polite     & polite    \\
		\bottomrule
	\end{tabular}
		\caption[Survey setup: Model output combinations (Part \num{2})]{
		Model combinations for the human evaluation to address whether the \gls{knnlm} outperforms the baseline \gls{lm}. Both models being compared use the style architecture.
	}
	\label{tbl:survey-output-pairs-part2}
\end{table*}

\begin{table*}
	\centering
	\small
	\begin{tabular}{lllll}
		\toprule
		\thead[l]{dataset} & \thead[l]{source style} & \thead[l]{datastore} & \thead[l]{target style} \\
		\midrule
		Formality   & formal    & formal         & formal      \\
		        &           & informal       & informal    \\
		        & informal  & formal         & formal      \\
		        &           & informal       & informal    \\
		        \midrule
		Toxicity     & neutral   & mixed          & non-toxic, toxic   \\
		        &           & non-toxic      & non-toxic   \\
		        &           & toxic          & toxic       \\
		        & non-toxic & mixed          & non-toxic, toxic   \\
		        &           & non-toxic      & non-toxic   \\
		        &           & toxic          & toxic       \\
		        \midrule
		Politeness     & impolite  & impolite       & impolite    \\
		        &           & mixed          & impolite, polite    \\
		        &           & polite         & polite      \\
		        & polite    & impolite       & impolite    \\
		        &           & mixed          & impolite, polite    \\
		        &           & polite         & polite      \\
		 \bottomrule
	\end{tabular}
		\caption[Survey setup: Model output combinations (Part \num{3})]{
		Model combinations for human evaluation to address whether the style architecture outperforms the baseline \gls{knnlm}.
	}
	\label{tbl:survey-output-pairs-part3}
\end{table*}

\end{document}